\documentclass[]{interact}

\usepackage{endnotes}
\usepackage{amsmath}
\usepackage{multirow}
\usepackage{pgfplots}
\usepackage{footnote}
\usepackage{comment}
\makesavenoteenv{tabular}
\makesavenoteenv{table}
\usepackage{booktabs}
\usepackage{multirow}
\usetikzlibrary{arrows, automata}
\usepackage[ruled,vlined]{algorithm2e}
\usepackage{hyperref}
\hypersetup{
    colorlinks=true,
    linkcolor=cyan,
    filecolor=magenta,      
    urlcolor=cyan,
    citecolor=red,
}
\usepackage[flushleft]{threeparttable}
\usepackage{epstopdf}
\usepackage{subfigure}

\theoremstyle{plain}

\theoremstyle{definition}

\theoremstyle{remark}

\begin{document}

\articletype{Research Article}

\title{Uncertainty Modelling in Risk-averse Supply Chain Systems Using Multi-objective Pareto Optimization}

\author{
\name{Heerok Banerjee\textsuperscript{a}, Dr. V. Ganapathy\textsuperscript{b} and Dr. V. M. Shenbagaraman\textsuperscript{c}}
\affil{\textsuperscript{a}Dept. of Mathematics and Computer Science, University of Antwerp, Belgium}
\affil{\textsuperscript{b}Dept. of Information Technology, SRM Institute of Science \& Technology, India}
\affil{\textsuperscript{c}Faculty of Management, SRM Institute of Science \& Technology, India}
\email{Heerok.Banerjee@student.uantwerpen.be\textsuperscript{a}}
}

\maketitle

\begin{abstract}
One of the arduous tasks in supply chain modelling is to build robust models against irregular variations. During the proliferation of time-series analyses and machine learning models, several modifications were proposed such as acceleration of the classical levenberg-marquardt algorithm, weight decaying and normalization, which introduced an algorithmic optimization approach to this problem. In this paper, we have introduced a novel methodology namely, Pareto Optimization to handle uncertainties and bound the entropy of such uncertainties by explicitly modelling them under some apriori assumptions. We have implemented Pareto Optimization using a genetic approach and compared the results with classical genetic algorithms and Mixed-Integer Linear Programming (MILP) models. Our results yields empirical evidence suggesting that Pareto Optimization can elude such non-deterministic errors and is a formal approach towards producing robust and reactive supply chain models.
\end{abstract}


\section{Introduction}

\begin{figure}[h!]
\centering
\subfigure[LSTM model without uncertainty model]{%
\resizebox*{6.8cm}{!}{\begin{tikzpicture}[] \label{subfig:LSTM}
\begin{axis}[
    width=\linewidth,
    xlabel=Time (in s),
    ylabel=Revenue generated (in \$),
    xmax=600,xmin=0,
    line width=0.5pt,
    mark size=0.5pt]
    \addplot+[smooth] table[y=Actual,col
    sep=comma]{csv/lstm.csv};
    \addlegendentry{Actual}
    \addplot+[smooth] table[y=LSTM_Model,col
    sep=comma]{csv/lstm.csv};
    \addlegendentry{LSTM Model}
    \end{axis}
\end{tikzpicture}}}\hspace{5pt}
\subfigure[NAR model with uncertainty model]{%
\resizebox*{6.8cm}{!}{
\begin{tikzpicture}[]\label{subfig:NAR}
\begin{axis}[
    width=\linewidth,
    xlabel=Time (in s),
    ylabel=Revenue generated (in \$),
    xmax=600,xmin=0,
    line width=0.5pt,
    mark size=0.5pt]
    \addplot+[smooth] table[y=Actual,col
    sep=comma]{csv/nar.csv};
    \addlegendentry{Actual}
    \addplot+[smooth] table[y=NAR_UM_Model,col
    sep=comma]{csv/nar.csv};
    \addlegendentry{NAR Model}
    \end{axis}
\end{tikzpicture}}}
\caption{Performance analysis of LSTM and NAR model for sequence-to-sequence prediction} \label{fig:SCM_compare}
\end{figure}
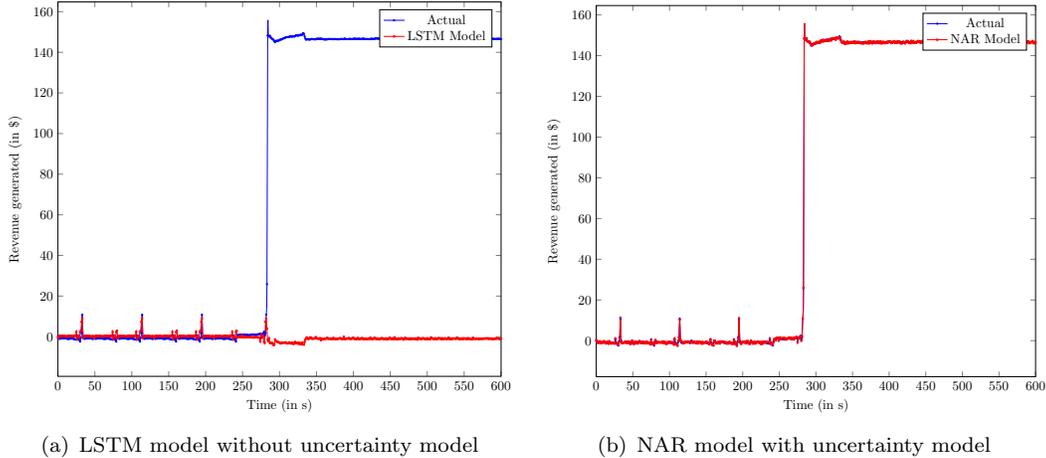

Time-series analyses have been widely employed in supply chain systems to support managerial strategies, predictive analytics and regression analysis. Although, time-series analyses is considered as a powerful tool applicable across a wide spectrum of objectives but careful modelling is still needed to improve accuracy and soundness. Contemporary Deep Neural Networks (DNNs) such as Long Short-Term Memory networks (LSTM), Recurrent Neural Networks (RNNs) and time-series models such as Non-linear Auto Regressive (NAR) are considered to be the state-of-the-art models for sequence-to-sequence prediction. However, empirical studies have shown that they perform poorly when there are induced variations in a time series sequence \cite{decaymodels}[Fig. \ref{fig:SCM_compare}]. Statistical analyses of supply chains suggest that decomposition techniques such as weighted normalization, discretization or quantization and transformations are feasible to handle variability, however it was insufficient to outline the performance trade-offs with time-series sequences under the influence of fuzzy environments\cite{TSperformance}. A later study conforming to dynamic analysis of time-series model was conducted and it was shown that Information Gain (IG) \footnote{Information Gain is a measure of entropy. It denotes the predictability of a variable w.r.t observations on previous timesteps.} was deterministic under certain assumptions\cite{uncertainty,TSperformance}. These insightful results played a central role in reinforcing some of the contemporary models but a key question, which still needs to be addressed is how to formally derive uncertainty models that can enhance existing time-series models with adaptive capabilities. In essence, the capability to learn invariants on previous time-steps can elude some of the non-determinism and it can parallely serve as a handy tool to produce reactive and error-tolerant models.

\par To further yield empirical evidence, we had previously modelled a supply chain system with periodic fluctuations in consumer demands \cite{dataset,SCMthesis}. Fig.\ref{subfig:LSTM} and Fig. \ref{subfig:NAR} plots the actual and the model predictions of supply chain revenues as a function of time. We clearly observe that the LSTM model is intolerant to induced variation and fails to predict accurately. Whereas, NAR model sufficiently adapts to periodic fluctuations and the error is minimal since the uncertainty model bounds the target sequence.

In this paper, we have introduced \texttt{Pareto Optimization} as a formal methodology to deal with uncertainty in dynamic multi-variate time-series sequences. In section \ref{background}, we have discussed the literature pertaining to optimization of supply chains systems and some of the recent contributions made in resolving to  accurate formal methods. To compare the proposed methodology with existing formal methods, a succinctly defined method namely \texttt{Mixed-Integer Linear Programming (MILP)} is selected and its fundamental limitations are discussed in section \ref{casestudy}. We have introduced mathematical models to represent objective functions and proposed a set of linear optimization scenarios based on a case study. To demonstrate its implementation, we have also provided numerical examples in section \ref{numAnalysis} and illustrated risk modelling in supply chain systems using MILP. In section \ref{Pareto}, we have demonstrated Pareto Optimization for multi-objective optimization problems using a genetic approach \cite{multiobjective} and pointed out its significance with respect to its methodological benefits. We have principally argued that the optimization scenario in MILP models is imprecise for certain corner cases where the optimization boundary lacks to accommodate trade-offs between multiple optimal solutions. We have shown that transforming linear optimization models into accommodating multi-objective scenarios can somehow elude those trade-offs. In section \ref{ParetoResults}, we have discussed the obtained results with visual aids and summarized the implementational complexities as compared to linear programming models. Finally in section \ref{conclusion}, we have concluded this study and provided insights to further improvements in Pareto optimization models for supply chain systems.

\section{Background \& Related Work}
\label{background}

Optimization problems in supply chain systems have attracted researchers and supply chain managers since the early industrialization and democratization of products at the expense of outsourcing services like manufacturing, dispensing, transportation, packaging etc. Optimum resource configuration is one of the most challenging problems in SCMS \cite{num7,num8}. For example, supply chain systems that are immune to pervasive risks, tend to emphasize on defining singleton optimization problems with either total costs, Return on Investment or transportation cost as objective functions. However, framing a single objective-based supply chain needs prior exposure and additional optimization (of influencing parameters) such as lead-time optimization, stock optimization, inventory-level optimization etc \cite{num9}. Sometimes conflicting objectives can cause disruptions in the supply chain; and therefore, a thorough study outlining the trade-offs between objective functions is necessary prior to mathematical modelling. This deviates the attention from supply chain modelling towards satisfying conflicting objectives, termed as multi-objective optimization\cite{num10}.

\subsection{Multi-objective Optimization}

\par Many novel multi-objective optimization methodologies have been proposed with limited constraints and succinct objectives. For example, A location-inventory problem was investigated using a multi-objective self-learning algorithm based on non-dominated sorting genetic algorithm II (NSGA II) in order to minimize the total cost of the supply chain and maximize the volume fill rate and the responsiveness level \cite{num11}. The study subjectively defined Genetic Algorithms (GA) [see appendix \ref{appendix:GA}] as the base for resolving optimization problems with partial accuracy. However, it was later indicated that forward supply chains tend to optimize targets such as supplier distributions, lead time, total costs etc. brutally at initial generations but had inexplicable impacts on successive generations. Subsequently in another study, another multi-objective optimization model using Fuzzy logic was derived for forward and reverse supply chain systems for minimizing total cost and the environmental impact \cite{num12}. The Fuzzy-based model consisted of two phases. The first phase converted the probability distribution of multi-objective scenarios into mixed-integer linear programming models and subsequently into an auxiliary crisp representation by denoting optimization scenarios as linear inequalities. The second phase centered on fuzzification of the subjected inequalities and defining rules to search for the non-dominant solution \cite{num13}. A key limitation of this methodology was that the underlying fuzzy inference system produced multiple non-dominant solutions, which created ambiguity.
\par Several models have also been proposed for multi-echelon supply chains (Suppliers, Distributors, Manufacturers, Retailers) with open-loops. For example, A multi-product with multi-period random distribution model was proposed to compute with predetermined goals for a multi-echelon supply chain network with high fluctuations in market demands and product prices. Similarly, a two-period multi-criteria fuzzy-based optimization approach was proposed where the central focus was to maximize the participant's expected revenues, average inventory levels along with providing robustness of selected objectives under demand uncertainties \cite{num14}. This methodology was heavily scrutinized and was later discarded because the method accounted two non-conflicting objectives to serve its purpose whereas conflicting objectives resulted in different solutions. Considering travelling distance and traveling time as objective functions, a self-learning multi-objective model was proposed namely fuzzy logic non-dominated sorting genetic algorithm II (FL-NSGA II), which basically solved a multi-objective MILP optimization scenario of allocating vehicles with multiple depots based on average traffic counts per unit time, average duration of parking and a primitive cost function \cite{num15}. This methodology reduced the operational complexity and upfront implementation time but the limitations remained the same as in \cite{num14}. Alternatively, a new hybrid methodology was also proposed which combined robust optimization (soft-computing), queuing theory and fuzzy inferential systems for multi-objective optimization based on hierarchical if-then rules \cite{num18}. Another variant of these hybrid soft-computing approaches consisted of modifying particle-swarm optimization method (MEDPSO) with classical genetic based approaches. These model were adopted for generating non-dominant solutions for a multi-echelon unbalanced supplier selection scenario \cite{num20,num21,num22}. 

By modifying classical GA-based optimization approaches, it was evident that hybrid approaches produced more dominant solutions in less time. However, empirical evidence suggested that these models were not accurate in terms of handling ripple propagations and demand-supply disruptions \cite{rippleEffect}. Arguably, it can be said that the emergence of the ripple effect redirected the attention of supply chain researcher towards emphasizing on modelling irregular variations, seasonal trends etc, which was coined as risks, in general. Gradually, the central focus which constituted of satisfying multiple objectives shifted towards bounding the entropy of the desired targets under the influence of risks.

\section{A case study of risk-averse supply chain systems } \label{casestudy}
Risks are indigenous across supply chains networks. This claim can be further solidified with some of the recent empirical studies conducted across dominant supply chain actors with respect to their logistics and operational strategies \cite{num6,num61,num7}. To reduce costs across the supply chain, large enterprises are moving manufacturing operations to countries which offer lower labor costs, lower taxes, and/or lower costs of transportation for raw materials. For some companies, outsourcing production involves not only a single country, but several countries for different parts of their products. Having suppliers in different geographic locations complicates the supply chain \cite{num13,num14}. Furthermore, companies have to  extend or maintain fast delivery lead times to customers who want to receive their products on schedule despite the increased complexity in the manufacturer’s supply chains \cite{num14}. Finally, companies also have to maintain real-time visibility into their production cycle ; from raw materials to finished goods and ensure the efficiency of their manufacturing processes. Due to these aforementioned complications, the inherent supply chain architecture induces risks that are abundant at every tier of the supply chain. Hence, in order to produce proactive managerial strategies and support business operations, building formal methods for modelling risks is a crucial step. \\
\par In the next section, one of the traditional approaches to modelling risks is discussed with a simple supply chain example. For the sake of the study, we have calculated the numerical methods for one iteration and we have assumed that the probability of failure for individual suppliers is zero.

\subsection{Modelling supply chains using Mixed-Integer Linear Programming (MILP)}
\label{MILP}
\begin{figure}[h!]
    \centering
    \begin{tikzpicture}[
            > = stealth, 
            shorten > = 3pt, 
            auto,
            node distance = 1.5cm, 
            thick 
        ]

        \tikzstyle{every state}=[
            draw = black,
            thick,
            fill = white,
            minimum size = 8mm
        ]
         \tikzstyle{elem}=[
            draw = black,
            thick,
            fill = white,
            minimum size = 4mm,
            node distance=2cm
        ]
        \tikzstyle{dem}=[
        node distance=2cm]
        
        \node[state] (S1) {$S_1$};
        \node[state] (S2) [below of=S1] {$S_2$};
        \node[state] (S3) [below of=S2] {$S_3$};
        \node[elem] (M) [right of=S2] {$M$};
        \node[elem] (R1) [right of=M] {$R$};
        \node[dem] (d) [right of=R1] {$demand$};

        \node[] (structS1) [left of=S1]{$\alpha_1, \beta_1$};
        \node[] (structS2) [left of=S2]{$\alpha_2, \beta_2$};
        \node[] (structS3) [left of=S3]{$\alpha_3, \beta_3$};


        \path[<-] (S1) edge node {$\$2$} (M);
        \path[<-] (S2) edge node {$\$3$} (M);
        \path[<-] (S3) edge node {$\$7$} (M);
        \path[<-] (M) edge node {$\$4$} (R1);
        \path[->] (d) edge node {} (R1);

\end{tikzpicture}
    \caption{Simple Supply chain architecture with multiple suppliers, single manufacturer and single retailer}
    \label{fig:example_arch}
\end{figure}
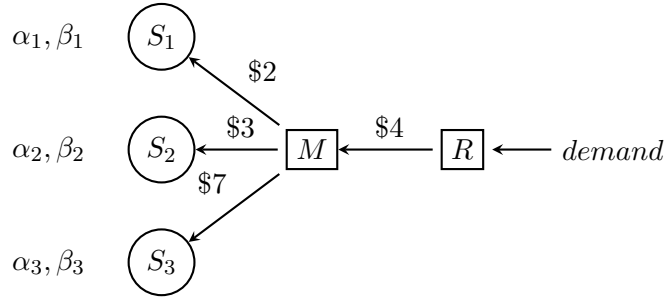

For a given supply chain with 3 echelons (tiers): \texttt{Supplier, Manufacturer, and Retailer} with selective activities such as sourcing or supplying raw material to each component, assembling final products and delivering to destination markets, each component has its own cost, lead-time and associated risk. Therefore, we can consider a simple supply chain network  as a directed graph G comprising of suppliers $S=\{ S_1,S_2,S_3\}$, a manufacturer $M$ and a retailer $R$ as shown in Fig.\ref{fig:example_arch}. For the sake of study, one must assume that the supply chain actors are independent and consequently the logistics and operational costs can be arbitrarily chosen.

In the above supply chain architecture, the demand subjected to the retailer is propagated directly to the manufacturer. Consequently, the demand is arbitrarily distributed among the suppliers $S=\{ S_1, S_2. S_3\}$. For example, if the demand is 'd' and the demand distribution is denoted by random variables $\{x_1, x_2, x_3\}$, then the cost function $F_c$ can be represented as a linear Diophantine equation. This is shown below:

\begin{enumerate}
     \item The total supply chain cost can be calculated as \cite{num7}:
    \begin{equation} \label{equ:totalcost}
    Total Cost=\xi \times \sum_{i=1}^{N}  (~\mu_{i}~.~ \sum_{j=1}^{N_i} C_{ij} y_{ij} \tag{Z1} )
    \end{equation}
    where, N is the number of components,\\
    $\xi$ denotes the period of interest,\\
    $\mu_{i}$ is the average demand per unit time,\footnote{For the sake of simplicity, we assume the period of interest as one year and the average demand per cycle as one/unit time. More realistic values are typically extracted by random sampling of time-series data. }\\
    $C_{ij}$ is the cost of the $j^{th}$ resource option for the $i^{th}$ component,\\
    $y_{ij}$ is a variable denoting the amount that the $i^{th}$ component supplies as a participant for the $j^{th}$ resource option\\
    
    Using \ref{equ:totalcost}, we can represent the supply chain model given in Fig.\ref{fig:example_arch} as:
    \begin{align}
    F_c(S): 2 x_1 + 3 x_2 + 7 x_3   \leq 4 d \tag{Z2} \label{equ:costFunction}\\
     s.t ~~~~x_1 + x_2 + x_3 = d \\
     0\leq x_1, x_2, x_3 \leq d 
\end{align}
    \par \footnotesize \color{red} Remark:\color{black}The inequality holds because we assume that the supply chain is in a supply-demand equilibrium. This is the case when the subjected demand has to be necessarily satisfied and the logistics costs for manufacturing, packaging etc is equally balanced by the cash flow throughout the supply chain.
    
    \normalsize
    \item The Risk Index is derived from the model proposed in \cite{risk_base}, and can be mathematically formulated as:
    
    \begin{equation}\tag{Z3}\label{equ:riskfactor}
    RI_{supplier}= \sum_{i,j=1}^{n} {\alpha s_{ij}~.~ \beta s_{ij}~.~
    (1-(1-\prod_{j=1}^{m} P(\tilde{S}_{ij}))} ~~ 
    \end{equation}

    where, for the $j^{th}$ demand,\\ $\alpha s_{ij}$ is the consequence to the supply chain if the $i^{th}$ supplier fails,\\
    $\beta s_{ij}$ is the percentage of value added to the product by the $i^{th}$ supplier,\\
    $P(\tilde{S}_{ij})$ denotes the marginal probability that the $i^{th}$ supplier fails \\
    
\end{enumerate}
\subsection{Numerical Analysis} \label{numAnalysis}

In this section, we attempt to replicate the model given in Fig. \ref{fig:example_arch} and observe the induced uncertainties in supplier workload distribution, which can be considered as an optimization problem. We construct a tabulated statistics representing the associated risks and logistics of the suppliers. Using \ref{equ:riskfactor} and \ref{equ:totalcost}, we compute the objective functions. Next, we rank all the possible distributions with respect to their fitness values. Finally, the global optimum is selected as the optimal workload distribution.

\begin{table}[htpb!]
\caption{Analysis of supplier workload distribution as an optimization problem}
\begin{threeparttable}
\begin{tabular}{cccccccccccc}
\\
\multirow{2}{*}{\begin{tabular}[c]{@{}c@{}}Supplier\\ Distribution\end{tabular}} & \multicolumn{6}{c}{Structural Variables} & \multicolumn{3}{c}{Assigned Probabilities} & \multirow{2}{*}{\begin{tabular}[c]{@{}c@{}}Total \\ Cost\end{tabular}} & \multirow{2}{*}{\begin{tabular}[c]{@{}c@{}}Risk\\  Index\end{tabular}} \\ \cline{2-10}
                                                                                 & $\alpha1$    & $\alpha2$   & $\alpha3$   & $\beta1$   & $\beta2$   & $\beta3$   & P($S_1$)        & P($S_2$)        & P($S_3$)        &                                                                        &                                                                        \\ \hline
{[}0,25,75{]}                                                                    & 0.1   & 0.1  & 1.0  & 0.1  & 0.3  & 1.0  & 0            & 0.25         & 0.75         & 600                                                                    & 0.7575                                                                 \\
{[}4,26,70{]}                                                                    & 0.1   & 0.1  & 1.0  & 0.1  & 0.3  & 0.8  & 0.04         & 0.26         & 0.70         & 576                                                                    & 0.5682                                                                 \\
{[}8,27,65{]}                                                                    & 0.1   & 0.1  & 1.0  & 0.1  & 0.3  & 0.8  & 0.08         & 0.27         & 0.65         & 552                                                                    & 0.5361                                                                 \\
{[}12,28,60{]}                                                                   & 0.1   & 0.1  & 1.0  & 0.1  & 0.3  & 0.8  & 0.12         & 0.28         & 0.60         & 528                                                                    & 0.4896                                                                 \\
{[}16,29,55{]}                                                                   & 0.1   & 0.1  & 0.8  & 0.2  & 0.3  & 0.6  & 0.16         & 0.29         & 0.55         & 504                                                                    & 0.2759                                                                 \\
{[}20,30,50{]}                                                                   & 0.2   & 0.2  & 0.8  & 0.2  & 0.3  & 0.6  & 0.2          & 0.3          & 0.5          & 480                                                                    & 0.266                                                                  \\
{[}24,31,45{]}                                                                   & 0.2   & 0.2  & 0.5  & 0.3  & 0.3  & 0.5  & 0.24         & 0.31         & 0.45         & 456                                                                    & 0.1455                                                                 \\
{[}28,32,40{]}                                                                   & 0.2   & 0.2  & 0.5  & 0.3  & 0.4  & 0.5  & 0.28         & 0.32         & 0.4          & 432                                                                    & 0.1526                                                                 \\
\color{black}
{[}32,33,35{]}                                                                   & 0.3   & 0.3  & 0.4  & 0.4  & 0.4  & 0.4  & 0.32         & 0.33         & 0.35         & 408                                                                    & 0.134
                       \\
\color{red}            
{[}50,25,25{]}                                                                   & 0.3   & 0.2  & 0.2  & 0.6  & 0.3  & 0.3  & 0.5         & 0.25         & 0.25         & 350                                                                    & 0.1560
                      \\
{[}70,20,10{]}                                                                   & 0.6   & 0.2  & 0.1  & 0.8  & 0.3  & 0.1  & 0.7         & 0.2         & 0.1         & 270                                                                    & 0.3341
\end{tabular}
\begin{tablenotes}
            \item[1] \footnotesize The values of $\alpha$ and $\beta$ are extensively discussed in \cite{risk_base}. The probability of failure for each supplier is the binomial probability of 'n' successive failures, which is given by $Prob_{S_{i}}= \sum_{k=0}^{n}{{n}\choose{k}}{1- {distribution_{S_{i}}}/{demand}} \times {{n}\choose{n-k}}{distribution_{S_{i}}}/{demand} $. \color{black}{The probability of failure is assumed to be zero.}\color{black}
            \item[2] \footnotesize It should be noted that we consider arbitrary distributions with absolute differences of atleast 5. The optimization scenario is subject to further evaluation of the entire supplier distribution. However, We overlook those distributions due to limited space.
            
            \item[3] \footnotesize The optimal distribution (\color{red}shown in red \color{black}) is selected accounting both the objective functions \ref{equ:riskfactor} and \ref{equ:totalcost}. We take the weighted sum of both objective functions to compute the fitness values. $(W_1 . Z1 + W_2 . Z2 = fitness )$
    
\end{tablenotes}
\end{threeparttable}
\end{table}


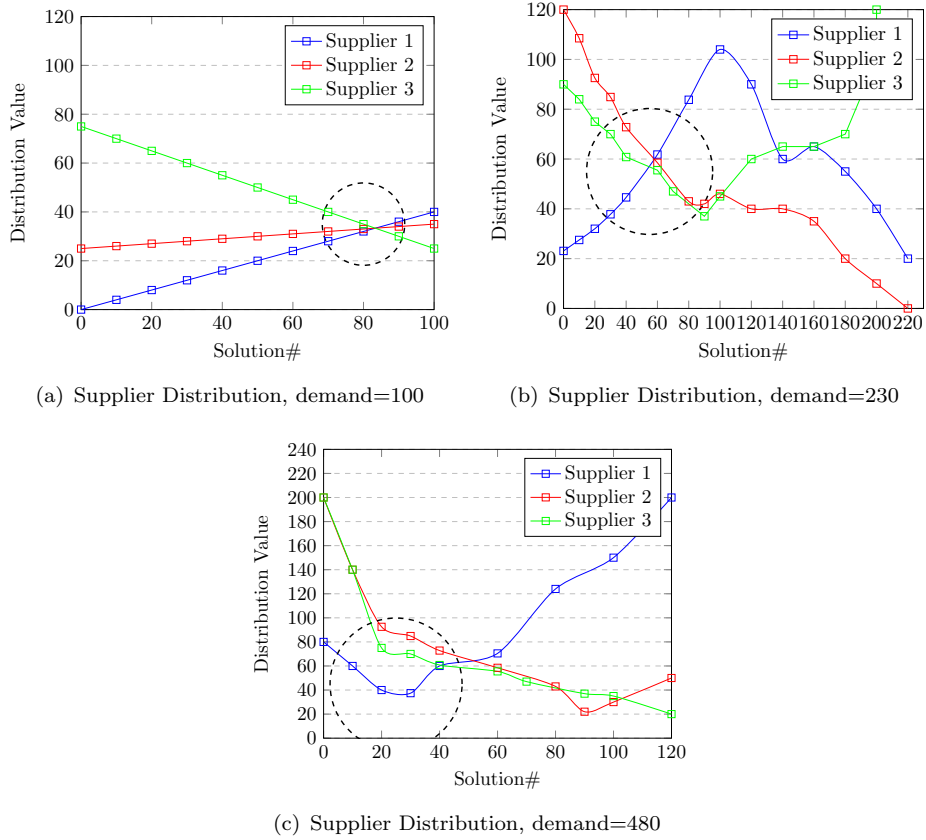
\begin{figure}[h!]
\centering
\subfigure[Supplier Distribution, demand=100]{%
\resizebox*{6cm}{!}{\begin{tikzpicture}[]

\begin{axis}[
    title={},
    xlabel={Solution\#},
    ylabel={Distribution Value},
    xmin=0, xmax=100,
    ymin=0, ymax=120,
    xtick={0,20,40,60,80,100},
    ytick={0,20,40,60,80,100,120},
    legend pos=north east,
    ymajorgrids=true,
    grid style=dashed,
]
 
\addplot[
    color=blue,
    mark=square,
    smooth
    ]
    coordinates {
    (0,0)(10,4)(20,8)(30,12)(40,16)(50,20)(60,24)(70,28)(80,32)(90,36)(100,40)
    };
    
    \addplot[
    color=red,
    mark=square,
    smooth
    ]
    coordinates {
    (0,25)(10,26)(20,27)(30,28)(40,29)(50,30)(60,31)(70,32)(80,33)(90,34)(100,35)
    };
    
    \addplot[
    color=green,
    mark=square,
    smooth
    ]
    coordinates {
    (0,75)(10,70)(20,65)(30,60)(40,55)(50,50)(60,45)(70,40)(80,35)(90,30)(100,25)
    };
    \draw[black,thick,dashed] (80,35) circle (0.8cm);

    \legend{Supplier 1, Supplier 2, Supplier 3,}
 
\end{axis}
\end{tikzpicture}}}\hspace{5pt}
\subfigure[Supplier Distribution, demand=230]{%
\resizebox*{6cm}{!}{
\begin{tikzpicture}[]
\begin{axis}[
    title={},
    xlabel={Solution\#},
    ylabel={Distribution Value},
    xmin=0, xmax=230,
    ymin=0, ymax=120,
    xtick={0,20,40,60,80,100,120,140,160,180,200,220},
    ytick={0,20,40,60,80,100,120},
    legend pos=north east,
    ymajorgrids=true,
    grid style=dashed,
]
 
\addplot[
    color=blue,
    mark=square,smooth
    ]
    coordinates {
    (0,23.1)(10,27.5)(20,32)(30,37.8)(40,44.6)(60,61.8)(80,83.8)(100,104)(120,90)(140,60)(160,65)(180,55)(200,40)(220,20)
    };
    
    \addplot[
    color=red,
    mark=square,smooth
    ]
    coordinates {
    (0,120)(10,108.5)(20,92.56)(30,84.9)(40,72.8)(60,58.5)(80,43)(90,42)(100,46)(120,40)(140,40)(160,35)(180,20)(200,10)(220,0)
    };
    
    \addplot[
    color=green,
    mark=square,
    ]
    coordinates {
    (0,90)(10,84)(20,75)(30,70)(40,60.8)(60,55.5)(70,47)(90,37)(100,45)(120,60)(140,65)(160,65)(180,70)(200,100)(200,120)
    };
    \draw[black,thick,dashed] (55,55) circle (1.2cm);

    \legend{Supplier 1, Supplier 2, Supplier 3,}
 
\end{axis}
\end{tikzpicture}}}
\subfigure[Supplier Distribution, demand=480]{%
\resizebox*{6cm}{!}{    \label{subfig:demand480}
\begin{tikzpicture}[]
\begin{axis}[
    title={},
    xlabel={Solution\#},
    ylabel={Distribution Value},
    xmin=0, xmax=120,
    ymin=0, ymax=240,
    xtick={0,20,40,60,80,100,120},
    ytick={0,20,40,60,80,100,120,140,160,180,200,220,240},
    legend pos=north east,
    ymajorgrids=true,
    grid style=dashed,
]
 
\addplot[
    color=blue,
    mark=square,smooth
    ]
    coordinates {
    (0,80)(10,60)(20,40)(30,37.4)(40,60)(60,70.45)(80,124)(100,150)(120,200)
    };
    
    \addplot[
    color=red,
    mark=square,smooth
    ]
    coordinates {
    (0,200)(10,140)(20,92.56)(30,84.9)(40,72.8)(60,58.5)(80,43)(90,22)(100,30)(120,50)
    };
    
    \addplot[
    color=green,
    mark=square,smooth
    ]
    coordinates {
    (0,200)(10,140)(20,75)(30,70)(40,60.8)(60,55.5)(70,47)(90,37)(100,35)(120,20)
    };
    \draw[black,thick,dashed] (25,45) circle (1.3cm);

    \legend{Supplier 1, Supplier 2, Supplier 3,}
 
\end{axis}
\end{tikzpicture}}}

\caption{Different supplier distributions under some degree of uncertainty} \label{fig:SupplierDistribution}
\end{figure}

The monotonous change in supplier distribution as we iterate through the solutions over x axis suggests that the maximum likelihood for an appropriate distribution is only concentrated where the demand is equally distributed to the suppliers. Intuitively, this can be assumed since larger share of workload on individual supplier will essentially increase unnecessary delays and processing latencies. The encircled region in Fig. \ref{fig:SupplierDistribution} denotes the minimum likelihood for risk-propagation and we attempt to bound the objective function \ref{equ:riskfactor} within the boundary of the enclosed circle. This assumption not only eliminates larger workload distributions but also gives a notion about the upper bound and the lower bound of the uncertainty being dealt with. This approach is convincingly applicable to most stationary supply chain systems where a constant demand is expected and the supplier distribution in convergent in nature. However, when the demand is increased, the risk distribution differs slightly. This is essentially due to the fact that larger demand queries would result in larger number of distributions, henceforth increasing the margin of the optimal solution. So, it would result in a different enclosed figure. In Fig. \ref{subfig:demand480}, since the distribution is skewed, there can be multiple distributions satisfying both the objective functions. In such cases, MILP fails to model uncertainties accurately as the optimal distribution maps to multiple risk indices. This essentially introduces ambiguity in the model, as the optimization problem transforms into a multi-objective optimization problem.

\section{Uncertainty Modelling using Pareto Optimization}
\label{Pareto}
\par \texttt{Pareto Optimal} solutions or \texttt{Pareto Fronts} are a set of trade-off points between different objective functions and comprises of a unique set of non-dominated solutions, that is unique global solutions for which no other solution exists that can improve one objective function without having to disrupt atleast one of the other objectives. In context to supply chain systems, we transform the linear optimization problem into a multi-objective optimization with some a-priori assumptions. We assume the upper bound and lower bound of both objective functions \ref{equ:riskfactor} and \ref{equ:totalcost} to be static over time. Then, the optimization scenario is accompanied with a genetic optimization approach, where a set of initial population, that is a set of arbitrary distributions are spawned and their respective fitness values are measured. Accordingly, a set of local minimas are calculated for a discrete value of the risk index. Mathematically, we attempt to map the Pareto optimal fronts with their corresponding risk indices and this can be formally represented as:

\begin{align}
     Objective Function: min/max.~ Total Cost  \tag{F1} \label{obj:cost} \\
     Objective Function:  min.~ \prod RI_{supplier} \tag{F2} \label{obj:risk}\\
     s.t,~~F_c(S): 2 x_1 + 3 x_2 + 7 x_3   \leq 4 d~~ holds~~ true \\
     0\leq x_1, x_2, x_3 \leq d
\end{align}

\begin{figure}[h!]
\centering
\subfigure[Maximum and Minimum Pareto Fronts, Demand=100]{%
\resizebox*{7.5cm}{!}{

\includegraphics[]{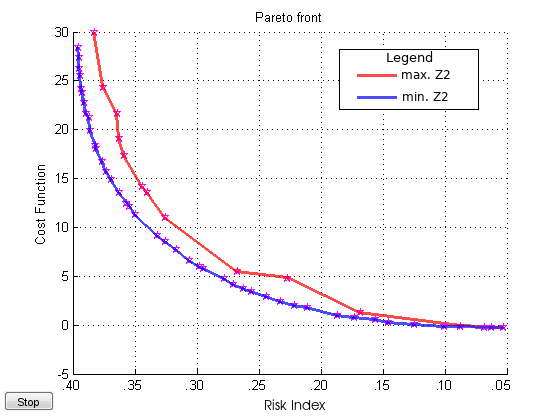}

}}\hspace{5pt}
\subfigure[Pareto Indicators]{%
\resizebox*{6.5cm}{!}{
\label{fig:paretoDistance}
\includegraphics[]{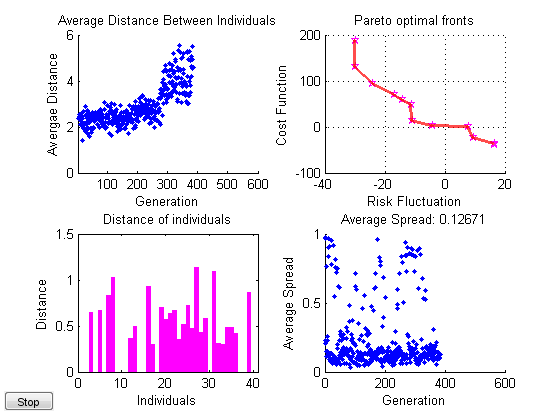}

}}
\caption{Pareto Optimization of objective functions \ref{obj:cost} and \ref{obj:cost}} \label{fig:paretofronts}
\end{figure}

For ease of computing, we can comfortably attempt Pareto optimization on the objectives \ref{obj:cost} and \ref{obj:risk} using MATLAB Optimization toolbox. In Fig. \ref{fig:paretofronts}, we have plotted the pareto optimal
fronts for a supply chain dataset \cite{dataset}. The curve color coded with \color{red}red \color{black} denotes the maximum boundary of the objective function \ref{obj:cost} with respect to objective \ref{obj:risk}. Likewise, the \color{blue} blue \color{black} curve denotes the minimum boundary of objective function \ref{obj:cost} with respect to objective \ref{obj:risk}. Finally, superimposing both curves yields a min-max domain of our cost function \ref{equ:totalcost}.   Fig. \ref{fig:paretoDistance} plots the average Pareto distances between successive generations.\footnote{The Pareto distance denotes the euclidean distance between two Pareto optimal fronts.} The average Pareto distance is quite satisfactory since it is as high as 1 unit upto 400 generations. A larger average Pareto distance would imply more data points to be interpolated in order to progress in our analysis and would increase the margin of interpolation errors. Fig. \ref{fig:uncertainFinal} illustrates the final plot that we desire. As observed, interpolated data points between adjacent Pareto fronts induce interpolation errors and inaccuracy of the model. Clearly, the number of Pareto fronts (and in turn, the initial population) plays a key role in determining the boundary of the target sequence accurately. To tackle this, a larger initial population can be set but at the cost of computing power.

\subsection{Implementation}
\label{GAstats}

In this section, we have summarized the implementation of the GA-based Pareto optimization approach in granular details. The implementation of the proposed methodology is carried out using \href{https://www.mathworks.com/products/optimization.html}{MATLAB Optimization Toolbox}. In \cite{dataset}, the supply chain network is simulated by explicitly modelling a Linear Diophantine Solver to compute the risk factor using \ref{equ:riskfactor}. A key limitation to this approach is that the solver selects the optimal supplier distribution based on the primitive objectives (total cost) only (equ.\ref{equ:costFunction}). In this paper, we have proposed a more accurate and sound methodology to account multiple objectives (risk factor, total cost, lead-time, ripple propagation etc.) and tackle optimization challenges with conflicting objective functions.

\par We employ a genetic approach for Pareto Optimization as described in algorithm \ref{algo:Pareto} and algorithm \ref{algo:GA}. Genetic Algorithms (GA) are a set of naturally inspired algorithms that are derived from Darwinian theory of evolution. Firstly, the desired objective functions are chosen (using \ref{equ:totalcost} and \ref{equ:riskfactor}) and a fitness function is assigned as per the supply chain model (using \ref{equ:costFunction}). Next, we chose an arbitrary demand and spawn random possible solutions. Subsequently, we evaluate the fitness function for each individual distribution and rank them according to their fitness values. At successive generations, we spawn further sets of randomly generated solutions and eliminate the unsuitable ones by comparing their fitness values. Finally, we rank the final surviving population at the last generation and the most dominant candidate, with the maximum fitness value is selected as the optimal solution.

\begin{algorithm}[ht!] \label{algo:Pareto}
\SetAlgoLined
\SetKwInOut{Input}{input}\SetKwInOut{Output}{output}
\SetKwFunction{GA}{Genetic Optimization}
\SetKwFunction{RI}{Risk Index}
\SetKwFunction{FC}{Cost Function}
\Output{ $Pareto Front=\{min.~ F_c,~ max.~ F_c,~ min.~ RI\}$ }
\Input{\{Cost Function: $2*x_1 + 3*x_2 + 7*x_3$,\\ Risk Index: \footnotesize{${{n}\choose{0}}{1- {distribution_S}/{demand}} \times {{n}\choose{n-0}}{distribution_S}/{demand}$},\\ \normalsize demand\}}

\Begin{
 $supplier_{min. cost}= \GA ~(Cost Function, demand, 0.8)$\;
 $supplier_{min. risk}= \GA ~(Risk Index, demand, 0.8)$\;
 \tcc{Add more objectives.}
 \BlankLine
 $min.~RI=  \RI ~(supplier_{min. risk})$\;
 $min.~F_c=  \FC ~(supplier_{min. cost})$\;
 $max.~F_c=  \FC ~(supplier_{min. risk})$\;
 \tcc{Add more targets.}
 \BlankLine
 \Return \{$min.~ F_c,~ max.~ F_c,~ min.~ RI$\}

}
 \caption{Multi-objective Pareto Optimization}
\end{algorithm}

\begin{algorithm}[ht!] \label{algo:GA}
\SetAlgoLined
\SetKwInOut{Input}{input}\SetKwInOut{Output}{output}
\SetKw{evalFitness}{evaluate fitness:}
\SetKw{spawn}{spawn :}
\Output{ S={$(d_1, d_2, d_3):~ d_1, d_2, d_3 \in N$} \tcc{optimal supplier distribution.}}
\Input{ \{Fitness function, demand, crossover\_ratio\}}
\SetKw{select}{random select :}
\SetKw{sort}{sort :}
\SetKw{nextgen}{next generation}

\Begin{
 \BlankLine
 \tcc{Initial population}
 \spawn random Initial\_Population\;
 \evalFitness Initial\_Population\;
 \sort Initial\_Population\;
 \BlankLine
 \tcc{Selection}
 set Final\_Population = Initial\_Population\;
 \While{$children_{eliminated}~!= 0$ or $lastGeneration == true$}{
  \spawn random children\;
  set $children_{eliminated} = zero$\;
  \For{each children}{
  \evalFitness children \;
  \eIf{$fitness_{children} \leq fitness_{Final\_Population[1]}$}{
  remove children\;
  set $children_{eliminated} += 1$\;
  }{
  set Final\_Population`=`children\;}
  }
  \BlankLine
  \tcc{Mutation}
  \For{crossover*100}{
    \select $children_1, children_2$ \;
    $children_3 = children_1 \oplus children_2$\;
    \BlankLine
    \evalFitness $children_3$\;
    \eIf{$fitness_{children_3} \leq fitness_{children_1} or fitness_{children_2} $}{
  remove $children_3$\;
  }{
  remove $children_1, children_2$\;}
  }
  \BlankLine
  \nextgen \tcc{repeat selection until last generation}
 }
 \BlankLine
 \tcc{Ranking}
 \sort Final\_Population\;

 \Return Final\_Population[1]\tcc{returns the most dominant solution w.r.t fitness function}
}
 \caption{Genetic Optimization Algorithm for supplier-selection problem}
\end{algorithm}

\pagebreak

\section{Results and Discussion}
\label{ParetoResults}
 In this section, we discuss some of the results obtained from the GA-based Pareto optimization approach in terms of performance, scalability and employability.

\begin{table}[ht!]
\caption{Population statistics for GA-based Pareto optimization, \textbf{demand=100}}
\footnotesize
\begin{tabular}{@{}ccccccc@{}}
\label{table:GAstats}

\toprule
\multirow{2}{*}{Generation}                                                      & \multicolumn{2}{c}{Simulation Variables} & \multicolumn{4}{c}{Simulation Statistics}                                                                                        \\ \cmidrule(l){2-7} 
                                                                                 & Parameter                 & Value        & Children             & Mutation             & Eliminated           & \begin{tabular}[c]{@{}c@{}}Total \\ Population\end{tabular} \\ \cmidrule(r){1-7}
\multirow{2}{*}{\begin{tabular}[c]{@{}c@{}}Generation 1\\ (Parent)\end{tabular}} & Initial Population        & 200          & \multirow{2}{*}{0}   & \multirow{2}{*}{160} & \multirow{2}{*}{179 \scriptsize{(89.5\%)}} & \multirow{2}{*}{21}                                         \\
                                                                                 & Cross-over ratio          & 0.8          &                      &                      &                      &                                                             \\ \\
\multirow{2}{*}{Generation 2}                                                    & Spawned Population        & 175          & \multirow{2}{*}{196} & \multirow{2}{*}{156} & \multirow{2}{*}{173 \scriptsize{(88.2\%)}} & \multirow{2}{*}{23}                                         \\
                                                                                 & Cross-over ratio          & 0.8          &                      &                      &                      &                                                             \\ \\
\multirow{2}{*}{Generation 3}                                                    & Spawned Population        & 240          & \multirow{2}{*}{263} & \multirow{2}{*}{210} & \multirow{2}{*}{238 \scriptsize{(90.4\%)}}  & \multirow{2}{*}{25}                                        \\
                                                                                 & Cross-over ratio          & 0.8          &                      &                      &                      &                                                             \\ \\
\multirow{2}{*}{Generation 4}                                                    & Spawned Population        & 87           & \multirow{2}{*}{112} & \multirow{2}{*}{89} & \multirow{2}{*}{85 \scriptsize{(75.8\%)}} & \multirow{2}{*}{27}                                        \\
                                                                                 & Cross-over ratio          & 0.8          &                      &                      &                      &                                                             \\ \cmidrule(lr){1-7}
\end{tabular}
\end{table}

\begin{figure}[ht!]
\centering
\begin{tikzpicture}[]
\begin{axis}[
    width=\linewidth,
    xlabel=Time (in s),
    ylabel=Revenue generated (in \$),
    xmax=250,xmin=0,
    ymax=16,
    line width=0.5pt,
    mark size=0.5pt]
    \addplot+[smooth] table[y=Actual,col
    sep=comma]{csv/nar.csv};
    \addlegendentry{Actual}
    \addplot+[smooth] table[y=NAR_UM_Model,col
    sep=comma]{csv/nar.csv};
    \addlegendentry{NAR Model}
    
    \addplot[
    color=green,
    mark=square,
    mark size=2pt,
    smooth
    ]
    coordinates {
    (0,0)(28,1)(32,13.7)(35,12)(38,0)(50,0.5)(60,0.2)(80,0)(100,-0.3)(110,1.1)(114,11.5)(120,-0.1)(140,0.3)(142,0.1)(150,0)(160,1)(163,-0.2)(180,-0.3)(188,-0.2)(203,-0.2)(208,-0.4)(210,0.4)(215,-0.2)(227,-0.2)(240,0.3)(250,2)
    };
    \addlegendentry{max. $F_c$}
    
    \addplot[
    color=black,
    mark=square,
    mark size=2pt,
    smooth
    ]
    coordinates {
    (0,-1)(5,-1.5)(15,-1.5)(30,-3)(38,-1.5)(50,-2)(60,-1.3)(80,-3)(100,-1.5)(110,-3)(114,-2)(120,-1)(140,-1.5)(142,-0.8)(150,-1.4)(160,-2)(163,-1.4)(180,-1.2)(188,-2)(203,-1)(208,-1.4)(210,-1.6)(215,-1.8)(227,-2)(240,-2)(250,-0.4)
    };
    \addlegendentry{min. $F_c$}
    \end{axis}
\end{tikzpicture}
\caption{NAR with superimposed Pareto optimal fronts \cite{dataset} \newline Initial Population=400, Number of obtained Pareto Fronts=27, Demand= 100 } \label{fig:uncertainFinal}
\end{figure}
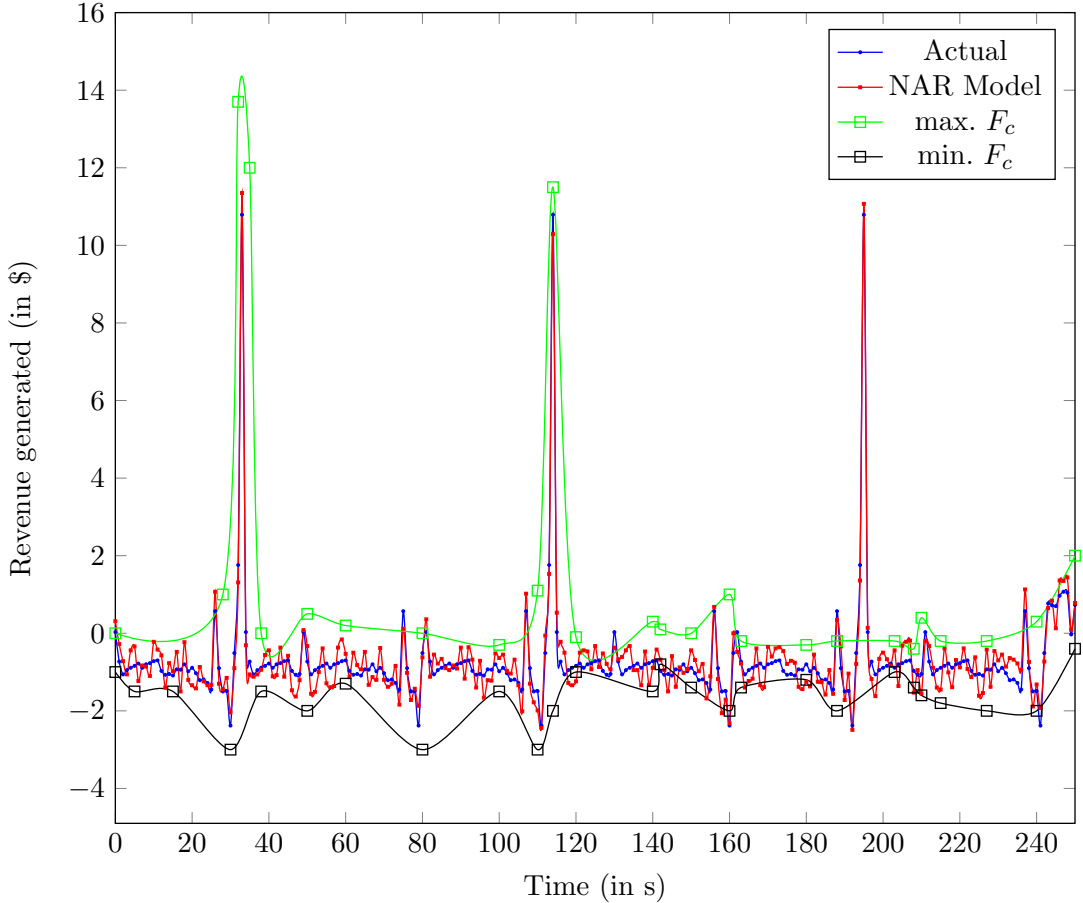

Table \ref{table:GAstats} provides a snippet of the population statistics while performing the GA-based Pareto optimization. We observe that the average elimination ratio for four successive generations is \texttt{85.9 \%}. This empirical evidence is crucial as it suggests that majority of the local optimum are eliminated at every successive generation. As opposed to linear optimization approaches such as MILP or dynamic programming, the time complexity in performing iterative comparisons is overwhelming, however, the statistics show that with a suitable cross-over ratio, most of the survived children represent the global optimum. It is also interesting to note that with a few obtained Pareto fronts, one can sketch the target boundaries for the objective function as shown in Fig. \ref{fig:uncertainFinal}. However, it is recommended to continue the genetic process for longer generations since hidden global solutions may be revealed at later generations.

\begin{figure}[h!]
\centering
{\begin{tikzpicture}[]

\begin{axis}[
    title={},
    xlabel={Demand},
    ylabel={Number of local solutions},
    xmin=0, xmax=1500,
    ymin=2000, ymax=350000,
    xtick={100,480,720,1200,1500},
    legend pos=outer north east,
    ymajorgrids=true,
    grid style=dashed,
]
 
\addplot[
    color=blue,
    mark=square,
    smooth
    ]
    coordinates {
    (100,2601)(240,14641)(480,49800)(720,82079)(1200,90000)(1500,251179)
    };
    
    \addplot[
    color=red,
    mark=square,
    smooth
    ]
    coordinates {
    (100,2800)(240,10847)(480,51240)(720,86780)(1200,116270)(1500,291761)
    };
    
    \addplot[
    color=green,
    mark=square,
    smooth
    ]
    coordinates {
    (100,2984)(240,17800)(480,58474)(720,88921)(1200,137899)(1500,341047)
    };

    \legend{MILP, classical GA, GA Pareto}
 
\end{axis}
\end{tikzpicture}}\hspace{5pt}
\caption{Comparison of MILP vs Classical GA vs GA-based Pareto Optimization}
\label{fig:comparisonOptimization}
\end{figure}
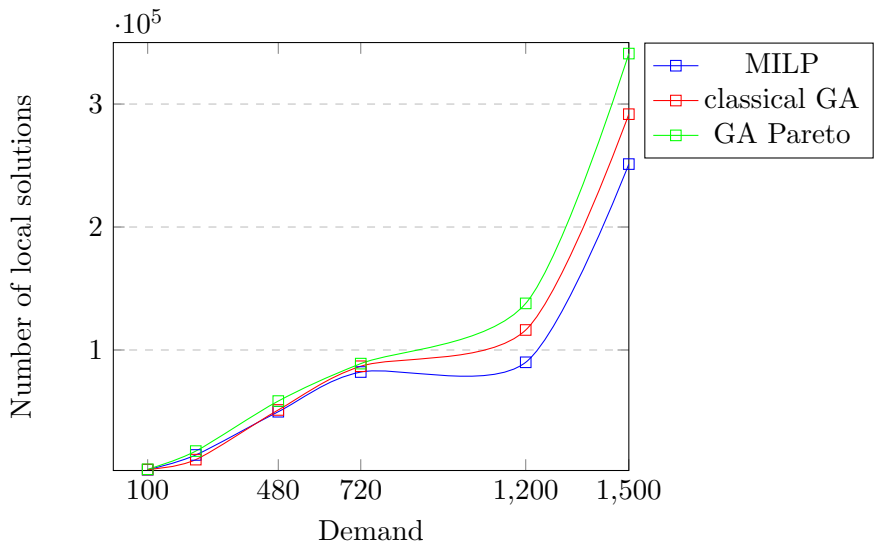

Next, we implemented the propesed GA-based Pareto Optimization for several demand values and compared the number of spawned solutions with MILP and classical GA-based optimization approaches. Fig. \ref{fig:comparisonOptimization} plots the results.
We observe that when the demand is high, which implicitly means that the number of possible solutions is quite large, MILP and classical GA-based approaches spawn quite lesser solutions as compared to GA-based Pareto optimization. This is essentially the case because MILP and classical GA methodology accompanies a sequential search and overlooks conflicting objectives. As such, this can be treated as a drawback of MILP and classical GA methodologies for resolving multi-objective optimization scenarios. However, the results indicate that essentially GA-based Pareto optimization produces hidden solutions that are overlooked by MILP models. Therefore, we can consider GA-based Pareto Optimization as a much dominant and sound approach as compared to MILP and classical GA methods.

Fig. \ref{fig:uncertainFinal} plots the time-series sequence with both minimum and maximum revenue generations, which are the obtained Pareto optimal fronts. We observe that by bounding the entropy of our objective function with respect to uncertain influencers, We essentially eliminate the need to mitigate the error and modify our forecasting models. As such, the ability to retain Information Gain at every timestep is crucial for time-series and machine learning models. We implemented the proposed GA-based Pareto Optimization with an Non-linear Autoregressive (NAR) model and observed that the forecasting model predicted real values within the obtained domain and with higher precision. This is our key insight for modelling uncertainties as well as non-deterministic influencers in supply chain systems.

\color{black}
\section{Conclusion}\label{conclusion}

In this study, we have discussed uncertainty modelling in time-series analyses with respect to domain-specific use cases such as multi-objective optimizations in supply chain systems. Firstly, we reviewed some of that empirical evidences that indicates to modify contemporary time-series models and assert uncertainty modelling as the root branch for time-series analyses. And we have discussed the employability of traditional approaches like Mixed-integer Linear Programming (MILP) along with mathematical models for examining risk-averse scenarios in supply chain systems. The results indicate that the uncertainties governing the operations of a supply chains are directly affected by the consumer demand, in the sense that increasing demand would reflect an increased risk fluctuation in the supply chain. Additionally, we argued that bulk demands would yield more possible distributions amongst suppliers, which would complicate our optimization scenario by increasing the solution space. We principally argued that if the number of possible distributions is strictly huge, it would result in numerous uneven distributions with variable risk indices. Our results indicated and supported that Pareto Optimization can essentially elude this trade-off and to some extent, mitigate the error induced in supply chain models. Finally, we implemented Pareto Optimization for a multi-tier supplier selection problem to optimize total cost and risk index. The empirical results indicated that the accuracy of the model is influenced by the initial population and the average Pareto distance of the model. However, we showed that even with random initial and candidate populations, a GA-based Pareto Optimization dominates over classical GA-based methods and traditional MILP methods.

\end{document}